\documentclass[10pt,twocolumn,letterpaper]{article}

\usepackage{iccv}
\usepackage{times}
\usepackage{epsfig}
\usepackage{graphicx}
\usepackage{amsmath}
\usepackage{amssymb}
\usepackage{kotex}
\usepackage{multirow}
\usepackage{caption}
\usepackage{subcaption}
\usepackage[dvipsnames]{xcolor}
\usepackage[breaklinks=true,bookmarks=false]{hyperref}

\newcommand\sh[1]{\textcolor{black}{#1}}
\newcommand\nj[1]{\textcolor{black}{#1}}
\newcommand\blfootnote[1]{%
  \begingroup
  \renewcommand\thefootnote{}\footnote{#1}%
  \addtocounter{footnote}{-1}%
  \endgroup
}

\iccvfinalcopy 

\def\iccvPaperID{****} 

\ificcvfinal\fi
\begin{document}

\title{URNet : User-Resizable Residual Networks with Conditional Gating Module}

\author{Sang-ho Lee$^{1,*}$, Simyung Chang$^{1,2,*}$, Nojun Kwak$^{1}$\\
$^{1}$Seoul National University, Seoul, Korea\\
$^{2}$Samsung Electronics, Suwon, Korea\\
{\tt\small \{shlee223, timelighter, nojunk\}@snu.ac.kr}
}
\maketitle

\begin{abstract}
   
Convolutional Neural Networks are widely used to process spatial scenes, but their computational cost is fixed and depends on the structure of the network used.
There are methods to reduce the cost by compressing networks or varying its computational path dynamically according to the input image. 
However, since \nj{a} user can not control the size of the learned model, it is difficult to respond dynamically if the amount of service requests suddenly increases.
We propose User-Resizable Residual Networks (URNet), which allows \nj{users} to adjust the scale of the network as needed during evaluation. URNet includes Conditional Gating Module (CGM) that determines the use of each residual block according to the input image and the desired scale. CGM is trained in a supervised manner using the newly proposed scale loss and its corresponding training methods. 
URNet can control the amount of computation according to user's demand without degrading the accuracy significantly. It can also be used as a general compression method by fixing the scale size during training. 
In the experiments on ImageNet, URNet based on ResNet-101 maintains the accuracy of the baseline even when resizing it to approximately 80\% of the original network, and demonstrates only about 1\% accuracy degradation when using about 65\% of the computation.
\end{abstract}

\section{Introduction}
\blfootnote{$^{*}$ Both authors contributed equally to this work.}
Generally, the computational graph in a deep neural network is fixed and unchanged during inference time. But in many situations of real applications, there may be the case that the system needs to handle various amounts of computation per request \cite{herbst2013elasticity}. For example, in the situation that the number of requests is rapidly increasing but the system is forced to respond quickly, it is better for the system to dynamically allocate less resource for requests within a moderate performance degradation bound. Or in the case that the environment's available amount of resource is varying by time to time, a fixed size of computation can be a waste of surplus resources.

Many researches working on compressing neural networks suggest architectures with less cost \cite{hinton2015distilling,howard2017mobilenets,iandola2016squeezenet}, but most of these architectures are fixed and static. Unlike these works, recent researches \cite{wu2018blockdrop,lin2017runtime,teerapittayanon2016branchynet} suggest the methods that a neural network dynamically changes its computation graph at test time, rather than fixed all the time. But these works only change the network path for each input, e.g., easy samples follow the path with less computation but complex samples require maximum available computation. 
\nj{Therefore,} these works \nj{can} not take care of the demand from the external environment. They are dynamic but cannot scale on our own purpose.

\begin{figure}[tb]
\begin{center}
   \includegraphics[width=1.\linewidth]{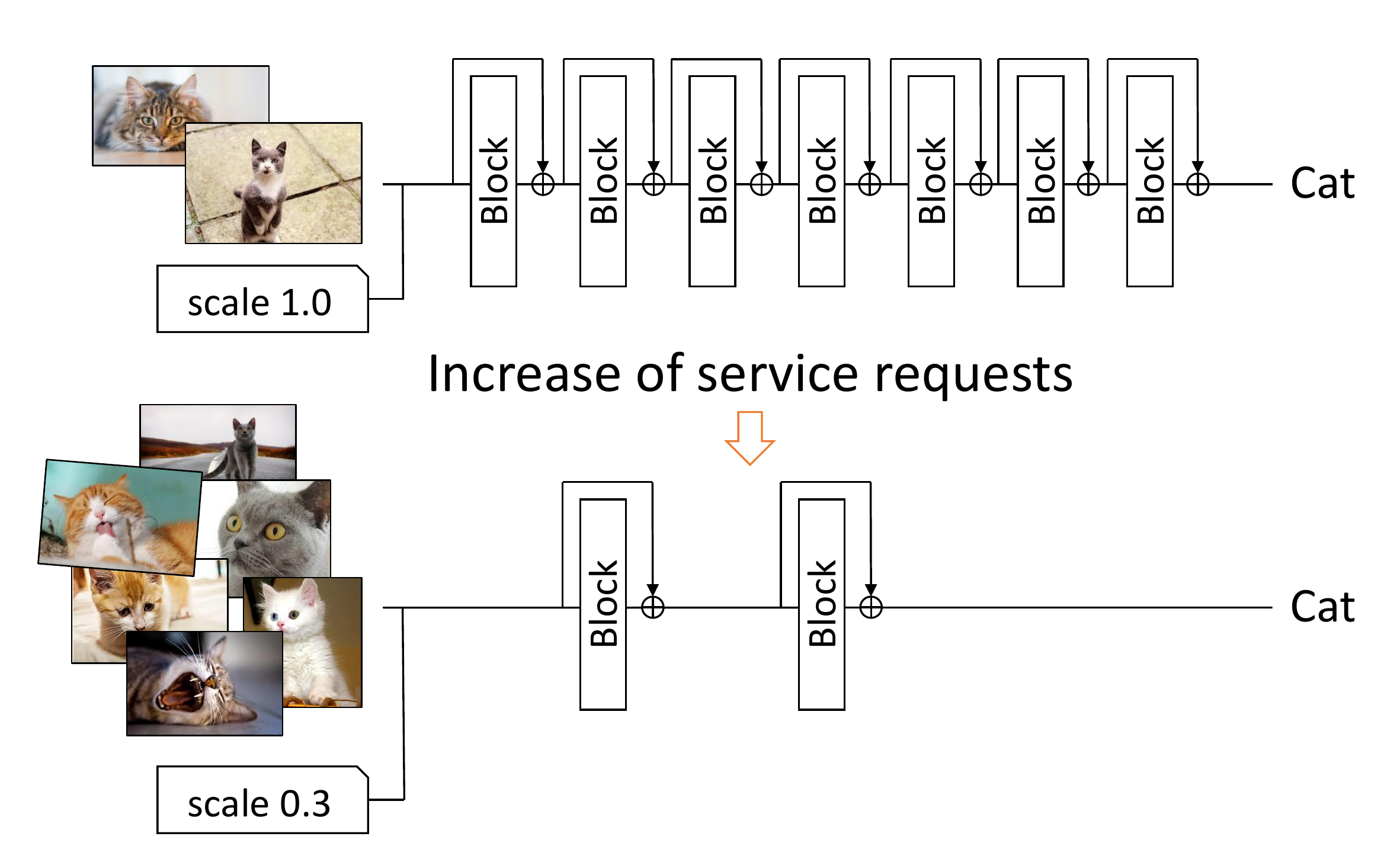}
\end{center}
\vskip -0.1in
   \caption{
\textbf{The concept of URNet.}
Our method uses the entire network when resources are sufficient. If the number of service requests increases, the system or a user can \nj{change} the scale of the network to use only \nj{a fraction} of \nj{the entire blocks}, thereby reducing the amount of computation in the network and processing the increased requests in time.
}
\label{fig:main_concept}
\end{figure}

\begin{figure*}[tp]
\centering
\includegraphics[width=0.98\linewidth]{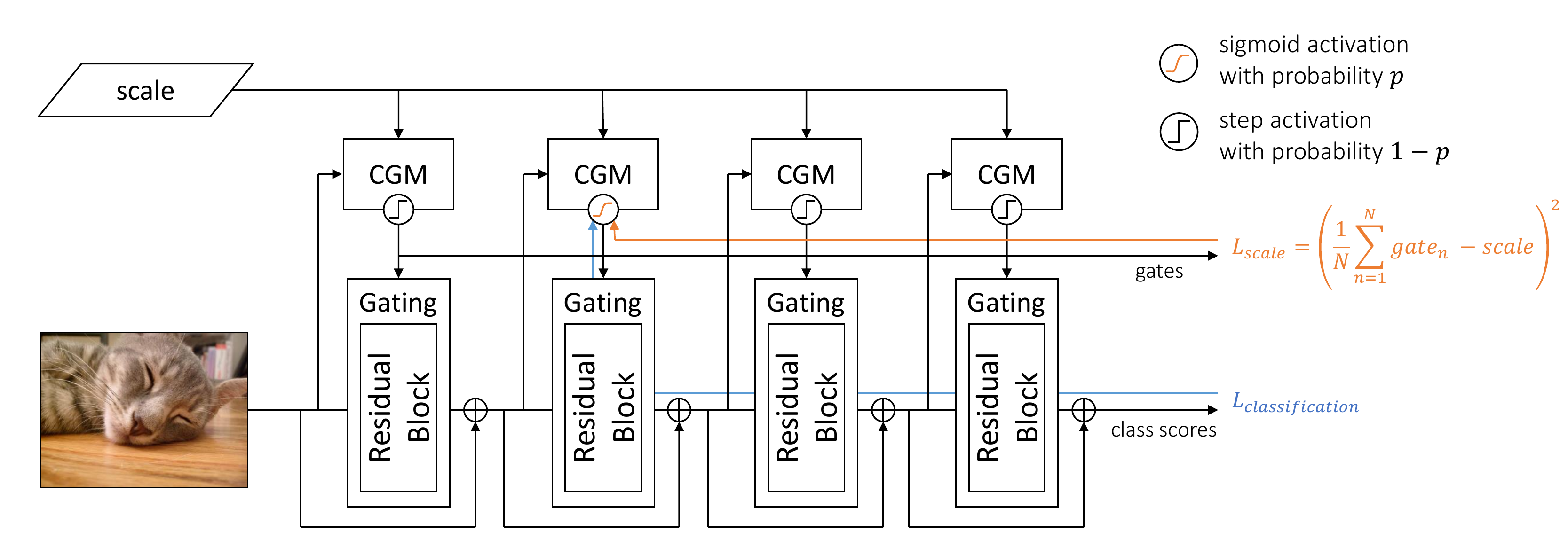}
\caption{\textbf{Overall structure of URNet.}
URNet locates Conditional Gating Module (CGM) for each residual block of ResNet, and determines whether each block is used or not. CGMs receive two inputs and output one gate value: one of the inputs is the features of the previous layer and the other is the desired scale parameter $\mathcal{S}$. To output a gate value, it uses a sigmoid function with probability $p$ and a binary step function with probability $1-p$ as an activation function. CGMs, with sigmoid function, can be trained through \textit{scale loss} $L_s$ and classification loss $L_c$ to increase classification performance while controlling the number of blocks used. At the time of inference, $p$ is set to 0, so that only the binary step is used and the amount of computation is reduced by not using the blocks with the gate value of 0.
}
\label{fig:overview}
\end{figure*}

In this paper, we suggest a model that can adjust its computational cost or \nj{the scale of a network} by itself, following given user's demand, like 70\% or 50\% of maximum resource for usage, at any inference time.
Our model is also variant to input samples, but its computational cost does not deviate significantly from the desired one.
It is \nj{robust} to the environment where the resources per request are limited or dynamically changing over time, and therefore, it fits such applications as in a backend server or background applications in a client.
Figure \ref{fig:main_concept} intuitively describes our concept. Our model is basically a plain ResNet \cite{he2016deep} architecture with additional gate modules located between neighboring blocks. Our gate module is computationally very cheap compared to the backbone network. \sh{Like the works in \cite{mirza2014conditional,sohn2015learning,chen2016infogan}}, these modules are conditional, that is, the \nj{user-specified scale condition of a network} can be fed into them. The network actually adjusts its scale by dropping some blocks of ResNet according to the binary output of the conditional gating module. As an experimental clue of feasibility that our method will not degrade much, the authors of \cite{veit2016residual} have suggested intuition that the ResNet is an ensemble combination of various subsets of residual blocks, and have experimentally shown that the ResNets are resilient to dropping layers. The work \cite{wu2018blockdrop} also has shown that the learned agent module can dynamically drop the layers even with performance gains.

Unlike \cite{wu2018blockdrop}, our conditional gating modules are trained without using reinforcement learning. To train the network so that it can gate the corresponding block, we may necessarily need to use 0/1 binary valued function which decides whether to use the component or not. Since this binary valued function is not differentiable, this problem \nj{has usually been} approached with reinforcement learning \cite{wu2018blockdrop,lin2017runtime}. On the other hand, we use a sigmoid function instead as a differentiable substitute of a binary function, and train the whole model while taking each gate module as either a sigmoid or a binary function based on a probabilistic rate. \nj{In addition, since} our training losses can be implemented from the conventional classification loss by just adding mean squared error loss between the desired input scale parameter $\mathcal{S}$ and the actual network scale, we do not need reinforcement learning, and the training is fast and stable.

We validate our model from the experiments on CIFAR-10, CIFAR-100~\cite{krizhevsky2009learning} and ImageNet~\cite{deng2009imagenet} datasets. By experiments, we show that our method can fit its scale \nj{to} a given condition well, and sometimes outperforms the baseline ResNet model when the scale \nj{parameter} $\mathcal{S}$ is 70\% or 80\%. Furthermore, even if we only use 60\% of blocks, the accuracy does not severely degrade.

Our contributions are summarized as follows:\\
(1) We propose URNet that can control the computational complexity of the model according to \nj{user's} demand.\\
(2) URNet does not suffer much from performance degradation even if it reduces the amount of computation.\\
(3) URNet is able to learn non-differentiable binary gates using a supervised learning method instead of using reinforcement learning, thus improving learning speed and stability.



\section{Related Works}
\label{sec:related}
\noindent\textbf{Model Compression}
There are many works on compression of neural networks. The network pruning \cite{luo2017thinet, he2017channel, he2018adc, yu2017nisp} removes the redundant channels of the network. The architecture search \cite{pham2018efficient,zoph2016neural} automatically finds \nj{an} efficient architecture by reinforcement learning. The works \cite{howard2017mobilenets,iandola2016squeezenet,chen2018big} are related to designing \nj{an} architecture of neural networks for more efficiency.
Most of these compression methods produce one \nj{static-sized} model. Using these models, to cope with the requirements of computing with limited resources, it should prepare many different-sized networks to its memory, which is not desirable for \nj{a resource-constrained} scenario like in an embedded environment.

\noindent\textbf{Dynamic Path Network}
The works in \cite{lin2017runtime,wu2018blockdrop,liu2017dynamic,odena2017changing,bolukbasi2017adaptive} are based on the idea of not fully using the network’s entire feed forward graph, but picking a subset of the graph specific for each input. Those methods have a full capacity network as a baseline, and train an additional module which selects where to forward between channels, blocks, or other paths. These selection modules can be other external small network \cite{lin2017runtime,wu2018blockdrop,liu2017recurrent}, or located inside the network as a gating function \cite{liu2017dynamic,bengio2015conditional,denoyer2014deep}. Utilizing this may lead the network to change its complexity at runtime. However, these methods usually need to be trained by reinforcement learning which normally encounters the problem of slowness and instability. Unlike these, our method is not based on reinforcement learning.



\section{User-Scalable Residual Networks}
Our goal is to train \nj{a} network to adjust its size according to the given \textit{\nj{desired} scale parameter $\mathcal{S}$} with a constraint of spontaneously minimizing the performance degradation. $\mathcal{S}$ can be any value between 0 and 1, representing what amount the user wants to scale the network. 
To achieve this, we propose \textit{User-Resizable Residual Networks} (URNet) and \nj{a} training method for them.
Figure \ref{fig:overview} is the overview of our URNet. We use ResNet as a baseline network, and drop the residual blocks spontaneously upon $\mathcal{S}$ to scale the network. 
For this, URNet includes a \textit{Conditional Gating Module} (CGM) as a method to decide whether to use each block \nj{or not}. This module is located one for each residual block, and outputs a gate value under the condition of input feature and the scale parameter $\mathcal{S}$.
To train the gates with the conventional supervised learning method, we propose a \textit{scale loss} and its training \nj{method}. More specifically, because binary gates can not be \nj{back-propagated} due to its non-differentiable characteristics, our gate has either a sigmoid or binary form at a certain probability during training. However, at the time of inference, it is always set as a binary step function.









\begin{figure}[tb]
\centering
\includegraphics[width=0.99\linewidth]{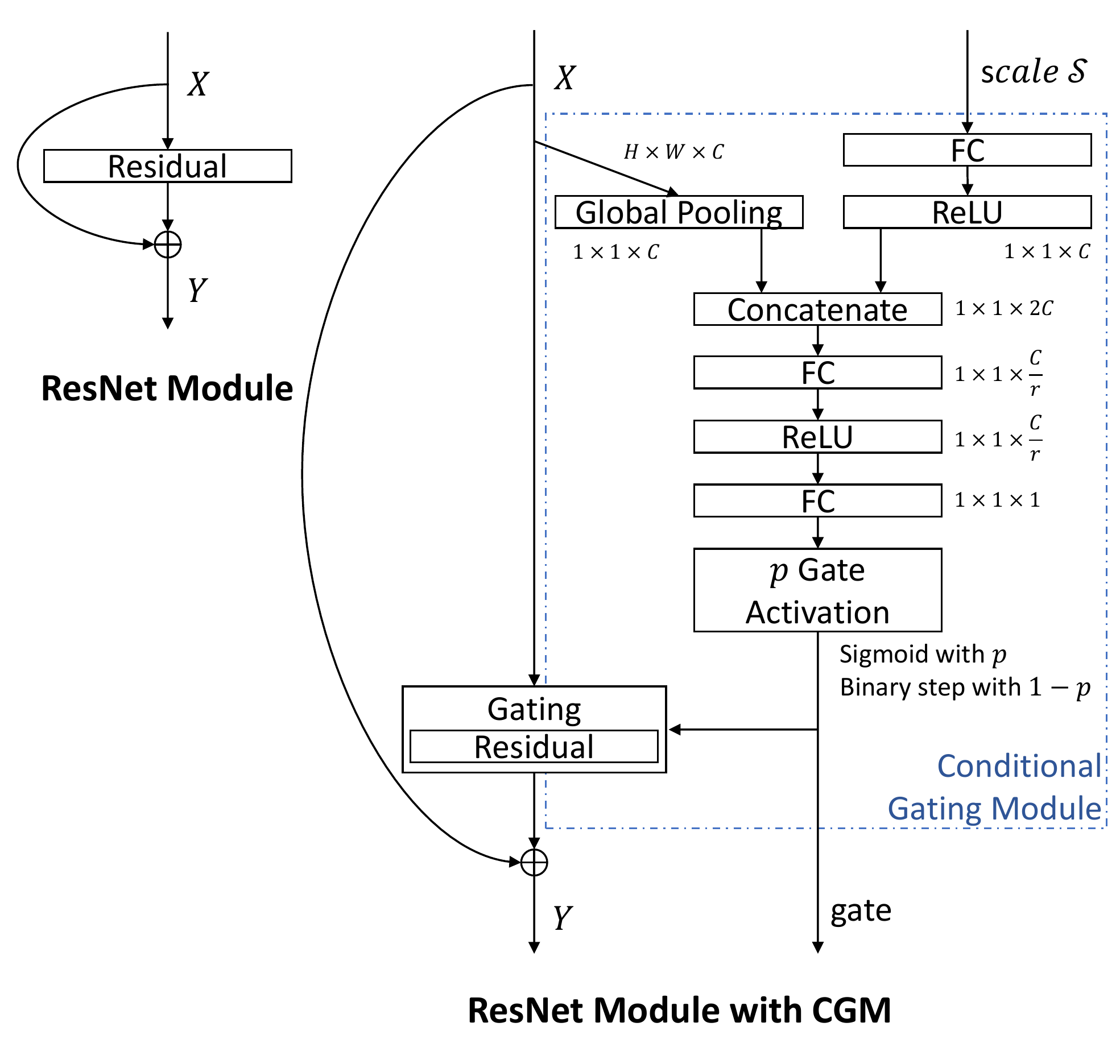}
\vskip -0.05in
\caption{\textbf{The structure of Conditional Gating Module (CGM).} The left is the ResNet module, and the right side is the ResNet module with CGM. 
CGM is a module that receives an input feature of residual block and a scale \nj{parameter} $\mathcal{S}$, then outputs a gate \nj{in} a sigmoid or binary form through a lightweight mapping function. 
}
\label{fig:CGM}
\end{figure}

\subsection{Conditional Gating Module}
\label{sec:CGM}
To determine whether to use each block of a ResNet, some modules or separate networks are required. And it must be decided before each block is activated. We propose a conditional gating module (CGM), a lightweight network module for this purpose. CGM is a simple structure that can be embedded in a ResNet, which can effectively determine whether to use a block or not with much fewer parameters and computation compared to those of the main ResNet.

As shown in Figure \ref{fig:CGM}, this gate module has two input entries, one is related to the features in the previous layer ($X \in \mathbb{R}^{H \times W \times C}$), and the other is the scale parameter $\mathcal{S}$. Thus, the gate module is conditional to both $X$ and $\mathcal{S}$. 
Several studies \cite{lin2013network,hu2018senet,chang2018broadcasting} have used global pooling to handle global features with fewer operations, and the proposed CGM also uses a global average pooling to handle the features of the previous layer. CGM then concatenates the globally pooled features of size $1 \times 1 \times C$ with $\mathcal{S}$ expanded to the same size, and outputs one gate variable through several fully connected layers and activation functions. In order to reduce the amount of computation, we use a reduction rate $r$ after the concatenation of the global feature and the scale condition like SENet \cite{hu2018senet}.


For scalability at inference time, we use \textit{Gate-Activation} function followed by several fully connected layers. 
\textit{Gate-Activation} is a simple function we designed for CGM, which works as either a sigmoid function or a binary step, depending on the given probability. CGM can learn only when the gate is sigmoid where gradients can be calculated and back-propagated. We call this frequency as a gate-training probability $p$. The Gate-Activation operates as a sigmoid function with probability $p$ and acts as a binary step function with probability $1-p$. 
In this case, training with binary gates plays a very important role as well. 
When all gates are activated with a sigmoid function, the remaining blocks cannot learn properly the cases of not using specific blocks. 
During the evaluation, this $p$ is fixed to 0 and only the binary function is used as an activation.

This gate module is then incorporated within the ResNet. Generally, the ResNet Module can be defined as:
\begin{equation}
Y= X + F(X),
\label{eq:res_out}
\end{equation}
where $X$ and $Y$ are the input and the output vectors of each layer, and the function $F(X)$ represents the residual mapping to be learned. The operation $X + F$ represents an element-wise addition as a shortcut connection.
Since the purpose of CGM is to gate the output of the function $F(X)$, the ResNet module with CGM can be \nj{expressed} as:
\begin{align}
\begin{split}\label{eq:CGM_sigmoid}
Y &= X + F_{gating}(F(X)) \\
&= X + CGM(X,\mathcal{S}) \cdot F(X) \\
&= X + gate \cdot F(X).
\end{split}
\end{align}
$F_{gating}$ refers to block-wise multiplication between $F(X)$ and a $gate$ which is the output of $CGM(X, \mathcal{S})$.
When the Gate-Activation works as a binary gate during evaluation, the module can be \nj{expressed} as:
\begin{equation}
Y=
\begin{cases}
X, & \text{if } gate = 0 \\
X + F(X), & \text{otherwise.}
\end{cases}
\label{eq:CGM_binary}
\end{equation}
As shown \nj{in} (\ref{eq:CGM_binary}), the corresponding block can be dropped if the value of the gate is 0. The computation of the block can then be reduced because $F(X)$ operation is omitted. 
For small $\mathcal{S}$, the gate values will have a good chance to be 0, depending on the input feature map $X$, resulting in a reduced computational complexity on average. \nj{On the other hand,} large $\mathcal{S}$ will mostly activate gates such that most blocks will be used for inference, resulting in high performance.

\subsection{Training CGM}
\label{sec:training}

\noindent \textbf{Scale Loss} 
In Section~\ref{sec:CGM}, it was mentioned that CGM can output \nj{a sigmoid or binary} gate, and can be learned through back propagation when using sigmoid gates. However, for actual learning, an objective function must be defined. The goal of our method is not \nj{only to} increase or maintain the performance of the classification, but \nj{also to} allow the user to change the size of the network \nj{according to} the desired one. Thus, the objective function must also satisfy both of these requirements. We propose a \textit{scale loss} that can be used with conventional supervised learning methods. This loss is defined so that the average of CGM gates is close to the scale parameter $\mathcal{S}$, as follows.
\begin{equation}
L_{s} = ((\frac{1}{N}\sum_{n=1}^{N}{gate_n})  - \mathcal{S})^2,
\label{eq:loss_scale}
\end{equation}
where $N$ denotes the number of residual blocks in the URNet and $gate_n$ represents the output of the CGM \nj{corresponding to} the $n$-th block. The full objective of URNet is the sum of this \textit{scale loss}, $L_s$, and the classification loss (cross entropy), $L_c$, of ResNet:
\begin{equation}
L = L_{c} + \beta L_{s}.
\label{eq:loss}
\end{equation}
Here, $\beta$ is a hyper parameter that controls the weights of $L_{c}$ and $ L_{s}$. Smaller $\beta$ means a bigger weight on classification, while bigger $\beta$ means a bigger weight on the scale loss. As $\beta$ increases, the actual block usage becomes similar to $\mathcal{S}$, but the classification performance may be sacrificed somewhat. Our experiments show that the number of actual blocks used can be controlled to be very close to the scale parameter $\mathcal{S}$.

\noindent \textbf{Gate Training Scheme} 
According to (\ref{eq:loss}), the CGMs are optimized to increase the classification performance and to make the average value of gates similar to the input parameter $\mathcal{S}$. However, in order to ensure that URNet operates at various values of $\mathcal{S}$ during inference, these values must be learned during training. This is done by 
randomly changing the range of $\mathcal{S}$ as we want to resize. 
The distribution of $\mathcal{S}$ is set as a uniform distribution of   ${\mathcal{U}}(\mathcal{S}_{min},\mathcal{S}_{max})$ during training. Here, $\mathcal{S}_{min}$ and $\mathcal{S}_{max}$ are the minimum and maximum of the range, respectively. Through this, the value of $\mathcal{S}$ and the actual block usage are synchronized with each other.

Since we use a pre-trained ResNet as the base network, we train only the CGM first, similar to BlockDrop~\cite{wu2018blockdrop} which trains the policy network first. This is to minimize the influence \nj{of premature CGM on the pre-trained ResNet}. After then, ResNet and CGM are jointly trained. 
However, by using the supervised learning method, CGM can be learned directly without using the method like the curriculum learning~\cite{bengio2013deep} which is used to overcome the instability of reinforcement learning in the BlockDrop paper, and learning can be performed with much less epochs. 
For CIFAR datasets, our method requires only 500 epochs which is a considerably smaller number compared to the training of BlockDrop which takes a total of 7,000 epochs including curriculum learning of 5,000 epochs.

\begin{table*}[t]
\caption{The accuracy (\%) and the number of block used under various scale conditions $\mathcal{S}$. The two row numbers in each cell are the accuracy (first row) and the number of blocks used (second row). Our method URNet(Ours) can be resized to match the user condition well, without severe accuracy degradation. Compared to the baseline with $\mathcal{S} = 1.0$ (93.2\% (CIFAR-10), 72.3\% (CIFAR-100)), our method performs \sh{better} for a wide range of $\mathcal{S}$ ($0.6 \sim 1.0$).}
\begin{center}
 \begin{tabular*}{0.99\textwidth} {@{\extracolsep{\fill} } p{0.15\textwidth}  c c c c c  | c c c c c}
 \hline
 								& \multicolumn{5}{c|}{CIFAR-10} 			& \multicolumn{5}{c}{CIFAR-100} 		        \\ 
 scale parameter $\mathcal{S}$ 	&  0.2	&  0.4	&  0.6	&  0.8	& 1.0  		& 0.2	& 0.4	& 0.6	& 0.8	& 1.0   		\\ \hline                                                      
 \multirow{2}{*}{
  \begin{tabular}[c]{@{}l@{}}ResNet-110\\
  (rand, val)\end{tabular}}
								& 11.8	& 15.4	& 28.0	& 68.6	& 93.2		& 1.3	& 2.4	& 7.3	& 38.1	& 72.3	 		\\ 
 								& 10.80	& 21.65	& 32.34	& 43.21	& 54.00		& 10.77	& 21.56	& 32.46	& 43.19	& 54.00			\\ \hline
 \multirow{2}{*}{
  \begin{tabular}[c]{@{}l@{}}ResNet-110\\
  (rand, train/val)\end{tabular}}
								& 83.3  & 91.0	& 92.8 	& 93.3	& 93.7		& 50.0	& 66.2	& 70.8	& 72.2	& 73.0	 		\\ 
 								& 10.80	& 21.65 & 32.38	& 43.16	& 54.00		& 10.83	& 21.58	& 32.40	& 43.24	& 54.00			\\ \hline \hline
 
 \multirow{2}{*}{
  \begin{tabular}[c]{@{}l@{}}External network\\
  (ResNet-8)\end{tabular}}
								& 91.5	& 92.6	& 92.7	& 93.1 	& 93.0	 	& 70.3 	& 71.1	& 71.4	& 72.5	& 72.5			\\ 
 								& 31.15	& 32.30	& 32.92	& 47.34	& 51.00		& 18.73	& 21.00	& 28.56	& 45.25	& 53.94			\\ \hline
 \multirow{2}{*}{
  \begin{tabular}[c]{@{}l@{}}URNet SG\\
  ($p=1.0$)\end{tabular}}
								& 12.7	& 21.4 	& 74.9 	& 81.7 	& 81.3		& 2.2	& 5.3	& 16.6	& 20.0	& 17.7	 		\\ 
 								& 6.75 	& 14.05	& 46.69	& 53.64	& 53.88		& 8.58	& 12.26	& 40.58	& 51.87	& 53.49			\\ \hline
 \multirow{2}{*}{
  \begin{tabular}[c]{@{}l@{}}URNet BG\\
  ($p=0.0$)\end{tabular}}
								& 93.2	& 93.1	& 93.0 	& 92.9 	& 92.8		& 71.5	& 71.6	& 71.7	& 71.8	& 71.7	 		\\ 
 								& 28.55	& 28.60	& 28.67	& 28.81	& 28.84		& 27.97	& 28.20	& 28.54	& 28.81	& 29.03			\\ \hline
 \multirow{2}{*}{
  \begin{tabular}[c]{@{}l@{}}
  ResNet+B/A\\
  (rand, train/val)
  \end{tabular}}
								& 83.1  & 91.1  & 92.5  & 93.3  & 93.7     & 50.1  & 66.4  & 70.4  & 72.2 & 73.2            \\
 								& 10.78 & 21.54 & 32.42 & 43.15 & 54.00	   & 10.79 & 21.56 & 32.44 & 43.22	& 54.00		    \\ \hline \hline
 \multirow{2}{*}{  	
  \begin{tabular}[c]{@{}l@{}}URNet(Ours)\\
  ($p=0.1$)\end{tabular}}
								& 92.2	& 93.3	& \textbf{93.7}	& \textbf{93.7} & 93.6	    & 70.7	& 71.5	& 72.4	& \textbf{73.0}	& 72.8			\\ 
 								& 18.08	& 20.86	& 32.02			& 44.37			& 52.19     & 28.10	& 28.57	& 32.00	& 44.61			& 49.41			\\ \hline
 \end{tabular*}
\label{table:cifar}
\end{center}
\end{table*}
\section{Experiments}

\subsection{Baselines and Experimental Setup}
In the following experiments, we have trained and evaluated our method on CIFAR-10, CIFAR-100 \cite{krizhevsky2009learning} and ImageNet \cite{deng2009imagenet} datasets. All accuracies we report are the top-1 accuracy. 
For ImageNet, we have evaluated the accuracy on the validation set of 50,000 images. For CIFAR-10 and CIFAR-100 we have tested on the test set of 10,000 images. The preprocessing of all dataset follows the code of  \cite{wu2018blockdrop}.
We have trained URNet from the pretrained ResNet model that the author of \cite{wu2018blockdrop} provided to the public. 
As a base network for our URNet, we have used ResNet-110 (54 blocks) for CIFAR datasets, and ResNet-101 (33 blocks) for ImageNet.
We have chosen the channel reduction rate $r$ of CGM (see Figure \ref{fig:CGM}) as 2 for CIFAR datasets and 16 for ImageNet. Similar to the evaluation of other compression methods, we calculate the number of multiply-accumulate operations of convolutional layers and linear layers in FLOPs (floating point operations). The total number of FLOPs of all the CGMs in ResNet-110 is only 0.04\% of the base network and 0.08\% for the ResNet-101. We train CGM only for 100 epochs on CIFAR datasets and 5 epochs on ImageNet. Then, we train CGM and the base network jointly for 400 additional epochs on CIFAR and 15 epochs on ImageNet. The learning rate is adjusted from $10^{-3}$ to $10^{-5}$.

\begin{figure}[t]
\centering
\includegraphics[width=1.0\linewidth]{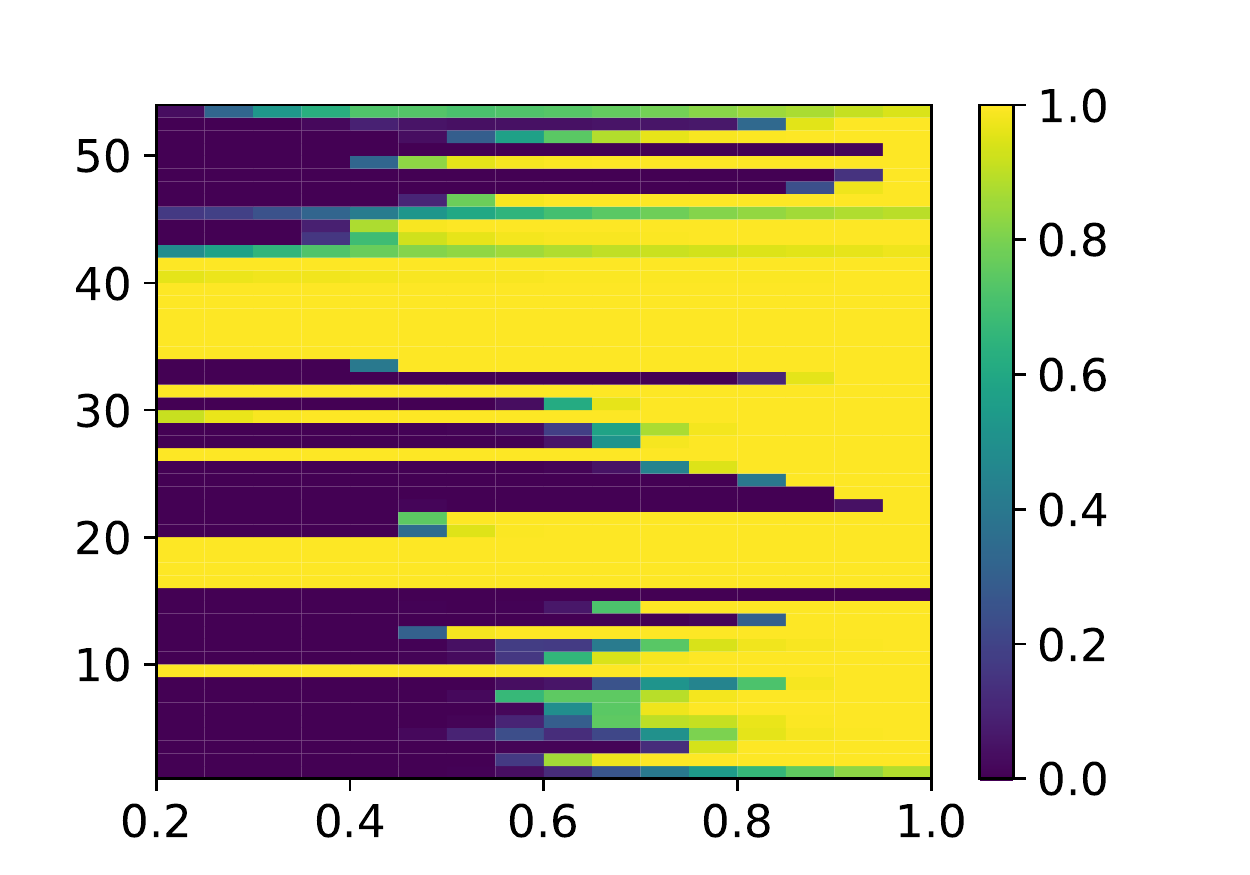}
\caption{The block usage map of URNet-110 on the CIFAR-10 test set with $\beta=4.0$. The horizontal axis is the scale parameter $\mathcal{S}$ and the vertical axis is the index of 54 residual blocks. As $\mathcal{S}$ increases, the usage of blocks gradually increases. Also, the presence of blocks whose usage is not 1 or 0, means that the usage of the block varies according to the input image even on the same scale.}
\label{fig:usage_map}
\end{figure}

\begin{figure*}[t]
\begin{subfigure}[b]{0.33\linewidth}
\includegraphics[width=\linewidth]{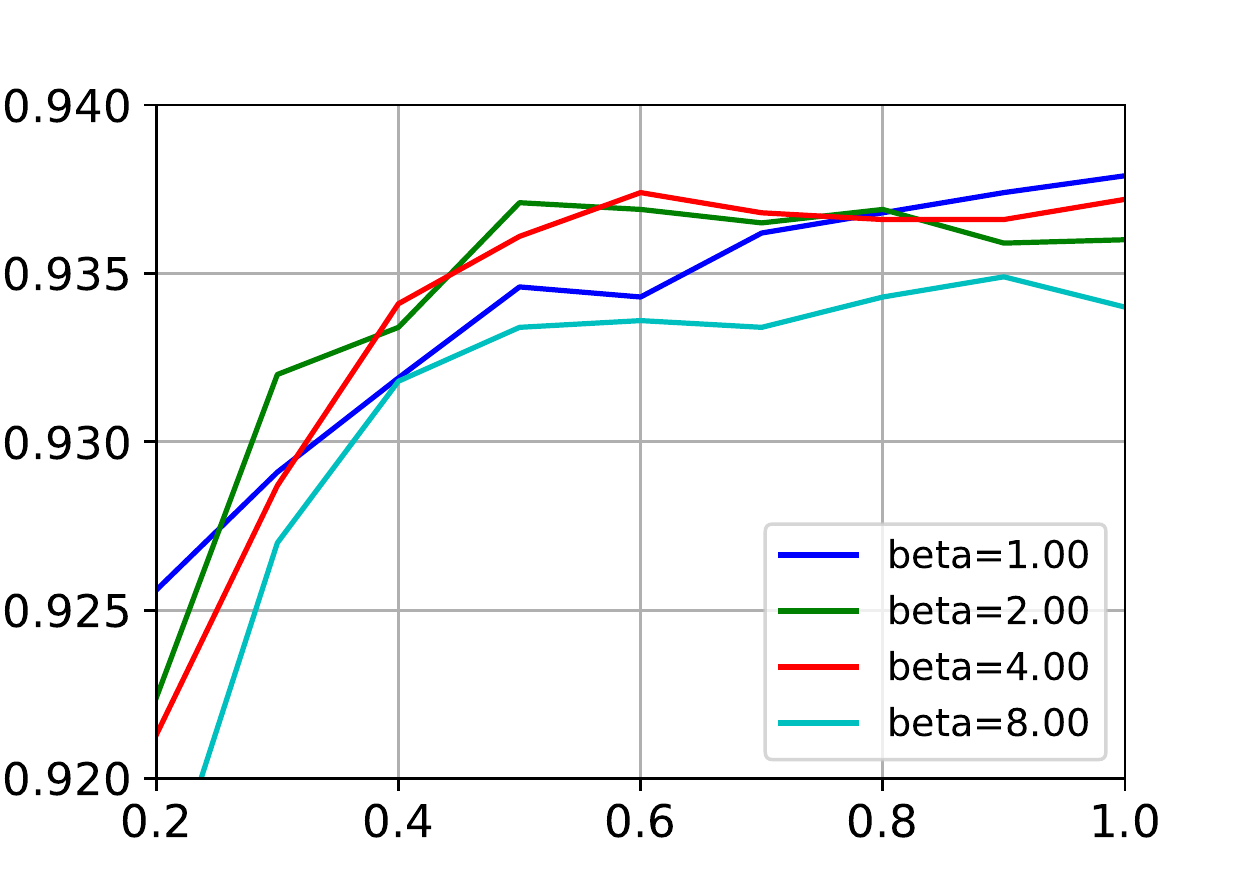}
\caption{Accuracy}
\end{subfigure} 
\begin{subfigure}[b]{0.33\linewidth}
\includegraphics[width=\linewidth]{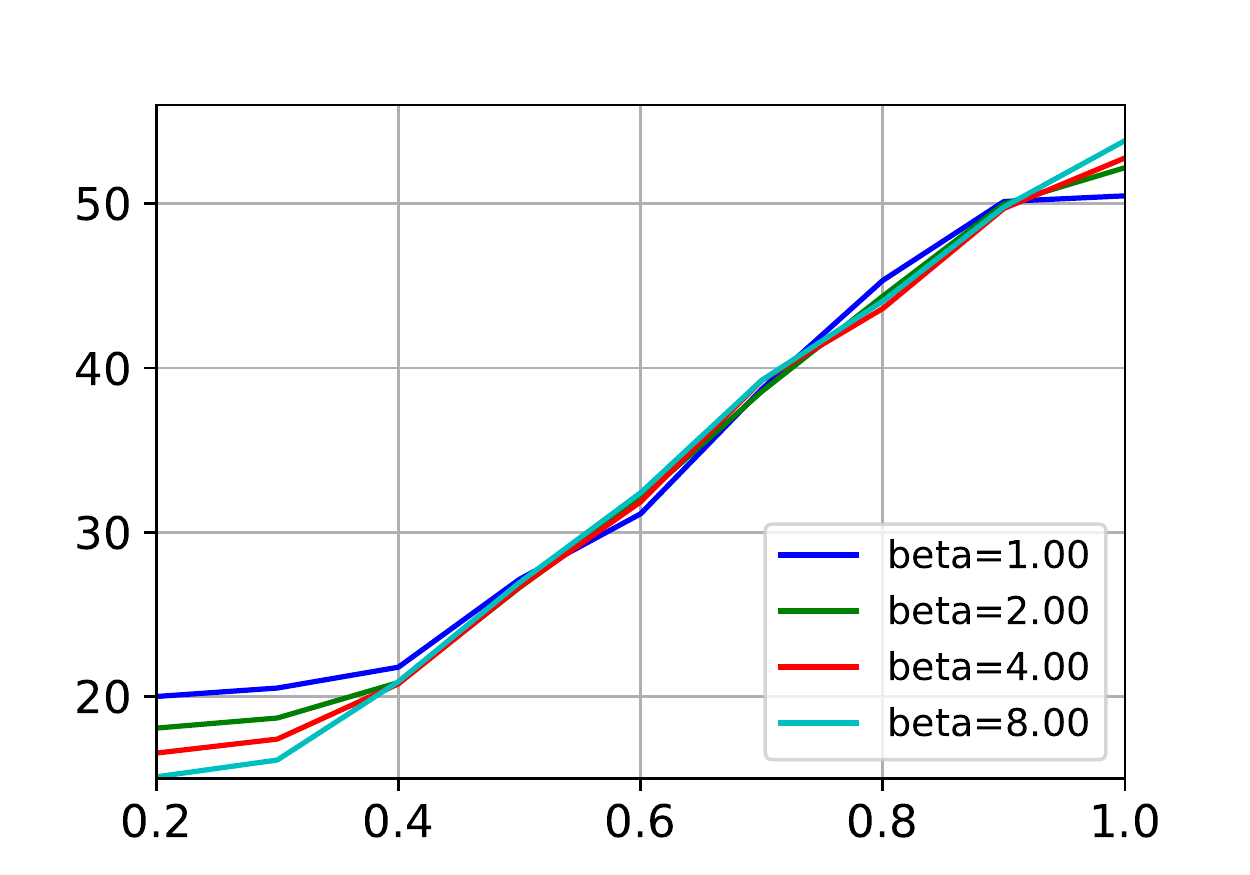}
\caption{Block usage}
\end{subfigure} 
\begin{subfigure}[b]{0.33\linewidth}
\includegraphics[width=\linewidth]{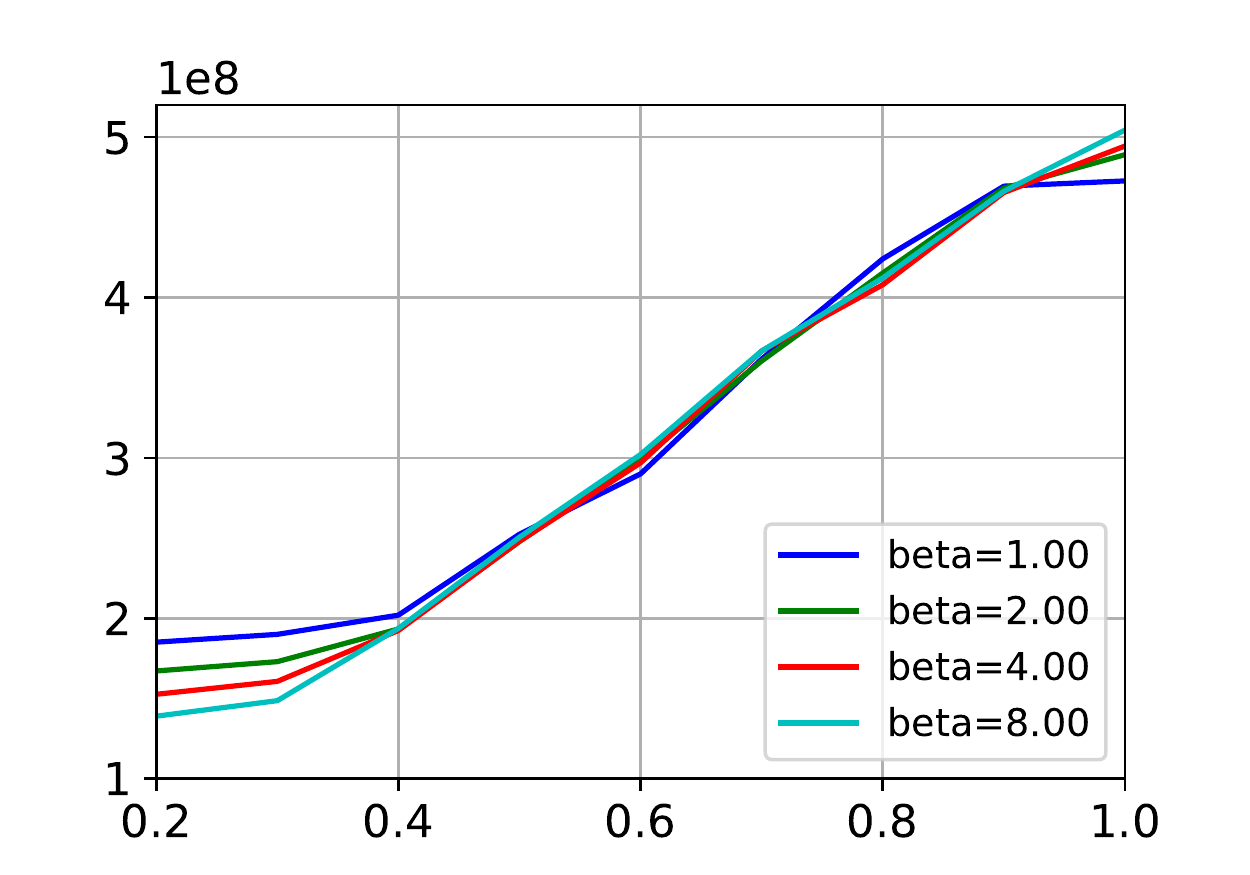}
\caption{FLOPs}
\end{subfigure} 
\caption{Accuracy, Block usage, and FLOPs versus scale parameter $\mathcal{S}$ under various $\beta$s, result of URNet-110 on CIFAR-10 dataset. The block usage and FLOPs follow the scale parameter $\mathcal{S}$ well, and better if $\beta$ is bigger. For accuracy, too big $\beta$ can downgrade the accuracy, so a moderate value of $\beta$ can perform better.}
\label{fig:beta}
\end{figure*}

\subsection{Result on CIFAR}
\label{sec:cifar}
Table \ref{table:cifar} shows the result of our method on CIFAR-10 and CIFAR-100, under various values of scale parameter $\mathcal{S}$. As shown in the table, our method can be resized as desired according to the given value of $\mathcal{S}$, without severe accuracy degradation. The table contains two baseline results of plain ResNet-110 which contains 54 residual blocks. It also contains the results of the proposed URNet (Ours), and other different settings with ablation. For those experiments we have set $\beta$ in equation (\ref{eq:loss}) as 2.0. During training, the scale parameter $\mathcal{S}$ has been uniformly sampled in the range of [0.2, 1.0], for every iteration.

The first and the second rows show the result of two baseline experiments with ResNet-110. The first row (ResNet-110 with rand, val) is the plain pretrained ResNet but we randomly drop the residual blocks at test time, to resize the network according to the given $\mathcal{S}$. The second row (ResNet-110 with rand, train/val) is the results of the finetuned ResNet that was trained with randomly dropping the blocks. It is not surprising that the performance of the second row is increased compared to the baseline at $\mathcal{S}=1.0$, because this can be interpreted as the dropout effect applied to block units. This result is very similar to the work in \cite{huang2016deep}, as they trained the ResNet with dropping each layer by a specific probability and unified them at test time. What is different from \cite{huang2016deep} is that they trained different drop probability for each blocks but ours is the same for all blocks.

The URNet (Ours) is trained with the gate training probability $p=0.1$, from the pretrained ResNet-110. Our method can match the network size to the desired value of $\mathcal{S}$ very well, without severe accuracy degradation. Note that at sizes in the range between 60\% and 100\% (32 blocks to 54 blocks), our method can even perform better than the baseline ResNet (93.2\% (CIFAR-10) and 72.3\% (CIFAR-100)). Unlike the finetuned dropped ResNet in the second row, our method does not severely degrade under very sparse block usage. 
Our method does not drop the blocks randomly like the compared method in the second row, but it drops the blocks by the decision of CGMs. This can be the reason for the lowered damage, as the CGMs can separate the blocks into most usable blocks and the remainder. As shown in Figure \ref{fig:usage_map}, under low $\mathcal{S}$ the CGMs have a tendency to open most important blocks exclusively, and these blocks are opened at every scale. And as $\mathcal{S}$ gets bigger, the rest of blocks gradually start to open (color changes from blue to yellow) because more blocks are getting more affordable.

\paragraph{Ablation Study}

In Table \ref{table:cifar}, there are \nj{4} 
other experiments for ablation study. The External network method uses an external small network with 3 residual blocks (equivalent to ResNet-8), which is separated from the base network, similar to the method presented in \cite{wu2018blockdrop}, but it is not trained using reinforcement learning. This external network is trained similar to the CGMs, by switching between sigmoid and binary activation with a rate of $p$. However, an important difference is that this module handles all of the gating at once with input data. It needs more computation compared to ours, but it is hard to expect them to extract rich features as it is smaller than the base network. As shown in the Table, the external network method does not work well to meet our purpose and the network usage deviates much from the scale parameter $\mathcal{S}$.

The URNet with sigmoid only (URNet-SG) is a special case of URNet with $p=1.0$, where the network is never trained with binary activation. But at inference time, all the CGMs are binary activated because our purpose is to drop some blocks. This experiment is a counter example that shows why the binary activation is needed during gate training. It shows that if we gate the block by just using a sigmoid value, the performance degrades severely. The URNet-BG is trained with $p=0$, which indicates that the network is trained with only binary activations. In this case, the CGMs are actually not trained and the block features are just multiplied by the untrained CGM output. The URNet-BG experiment shows that without sigmoid activation, the URNet can not resize to the desired \sh{size} $\mathcal{S}$ at all, and the result is just from an additional training (400 epochs) of an arbitrary subgraph of ResNet.

The ResNet+B/A always uses plain sigmoid function for the activation of CGM, which can be considered that $p=1.0$ at both train and test time. Note that the variants of URNet set $p=0.0$ at test time. In this case, it can learn the continuous block-wise attention (0$\sim$1), so possibly it gains more accuracy than the baseline ResNet.
However, the ResNet+B/A has no binary function, thus it should calculate all the blocks, which means that it is not resizable. Resizing it with a random drop during training (ResNet+B/A(rand, train/val)) results in similar performance with the second row of the Table.
It shows that the model can get accuracy gain with block attention, but suffers such a degradation when trying to resize by applying a random drop.
If we force the B/A module output to hard attention by thresholding the continuous attention at test time, it is identical to URNet-SG, which also fails to our purpose.
 Even if the CGMs in URNet does not utilize the gain from continuous block-attention (0$\sim$1), it outperforms the ResNet+B/A for most values of $\mathcal{S}$.
How the URNet does not suffer such degradation (and even gain accuracy) is that it can learn whether the block is necessary or can be abandoned, under given $\mathcal{S}$, by the proposed gate training scheme. As can be inferred from Figure \ref{fig:usage_map}.


\paragraph{\sh{Resize Ability}}

The hyper-parameter $\beta$ in (\ref{eq:loss}) can represent how strictly we want the network to follow the desired scale $\mathcal{S}$.
If we set $\beta$ higher, the network is more strongly affected by the scale loss. As shown in Figure \ref{fig:beta}(b), the higher $\beta$ becomes, the more strict the network becomes in following the target scale.
For lower $\beta$, the block usage is slowly fixed at the boundary of $\mathcal{S}$, especially when $\mathcal{S}=0.2$. If $\beta$ is too big, the accuracy of the network seems to be downgraded as shown in the case of $\beta=8.0$ in Figure \ref{fig:beta} (a). This is because too much scale loss can constrain the network capacity leading to a poor classification loss. But Figure \ref{fig:beta} (a) shows that $\beta$ and the accuracy does not have a complete negative correlation for relatively small $\beta$ ($\beta = 1,2,4$), and the maximum accuracy point lies between $\beta=1.0$ and $\beta=4.0$. Because our scale loss can work like regularization of the weight, under the proper choice of $\beta$, the network accuracy can be increased.


\subsection{Result on ImageNet}
\label{sec:imagenet}
Table \ref{table:imagenet} is our result on ImageNet (ILSVRC2012).
We trained the URNet from ResNet-101 which has total 33 blocks.
The result of ResNet-\{72, 75, 84, 101\} are brought from \cite{wu2018blockdrop}.
In this experiment, $\beta$ is set to 4.0.
Our method performs better than ResNets with the same amount of computation in all the cases.
When $\mathcal{S}$ is about 0.72, our URNet performs equal to ResNet-101 (accuracy: 76.4\%) using about 1.24E+10 FLOPs. The accuracy keeps increasing gradually with $\mathcal{S}$, and our best accuracy 76.9\% is achieved at $\mathcal{S}>0.95$. Note that the accuracy of ResNet-101+B/A(rand, train/val) is \{26.2\%, 49.9\%, 64.1\%, 71.6\%, 76.0\%\} for each $\mathcal{S}$=\{0.2, 0.4, 0.6, 0.8, 1.0\} (see B/A module in Ablation Study section).

\begin{table}[t]
\caption{The accuracy and the block usage under various scale condition. The baseline accuracy of ResNet-101 on ImageNet is 76.4. Our best accuracy is achieved at $\mathcal{S}>0.95$, which is 76.9\%.}
\vspace{-4mm}
\begin{center}
 \begin{tabular*}{\linewidth} {@{\extracolsep{\fill} } p{0.1\textwidth} c c c c}
 \hline
\multicolumn{4}{c}{ImageNet} 							\\ \hline
                                & \#Blocks			& {\small FLOPs(E+10)}	& Accuracy		\\ \hline
 ResNet-72                      & 24.0					& 1.17					& 75.8			\\
 ResNet-75                      & 25.0					& 1.21					& 75.9			\\
 ResNet-84                      & 28.0					& 1.34					& 76.1			\\
 ResNet-101						& 33.0					& 1.56					& 76.4			\\ \hline\hline
 \end{tabular*}
 \begin{tabular*}{\linewidth} {@{\extracolsep{\fill} } p{0.1\textwidth} c c c c c}
 $\mathcal{S}$ 					& 0.2		& 0.4		& 0.6		& 0.8		& 1.0			\\ \hline               
 Accuracy						& 74.0		& 74.9		& 75.7		& 76.4		& \textbf{76.9}	\\ 
 {\small Block usage}			& 18.78		& 19.77		& 22.01		& 26.94		& 32.00			\\ 
 {\small FLOPs(E+10)}			& 0.94		& 0.98		& 1.08		& 1.30		& 1.52			\\ \hline
 \end{tabular*}
\label{table:imagenet}
\end{center}
\vspace{-4mm}
\end{table}

\subsection{Qualitative Results}

Our CGMs can dynamically select the blocks to match the given scale parameter $\mathcal{S}$. But not only considering the given $\mathcal{S}$, the CGMs also consider the input features from the previous layer to decide whether to use the corresponding block or not. If the CGMs have not received information from the previous block, i.e., if we have trained the CGMs with using only $\mathcal{S}$ as an input, the goal of CGMs would be only to match the given scale parameter $\mathcal{S}$, resulting in the CGM's parameters to be fixed immediately from the start of the training. In this case, the URNet for a given $\mathcal{S}$ will be just some arbitrary fixed subgraph of baseline ResNet, whose size is $\mathcal{S}$. As Table \ref{table:woblock} shows, in this case, the variance of gate usage of CGMs without using the features from the previous block is absolutely zero.

\begin{figure}[t]
\centering
\begin{subfigure}[b]{0.48\linewidth}
\centering
\includegraphics[width=0.48\linewidth]{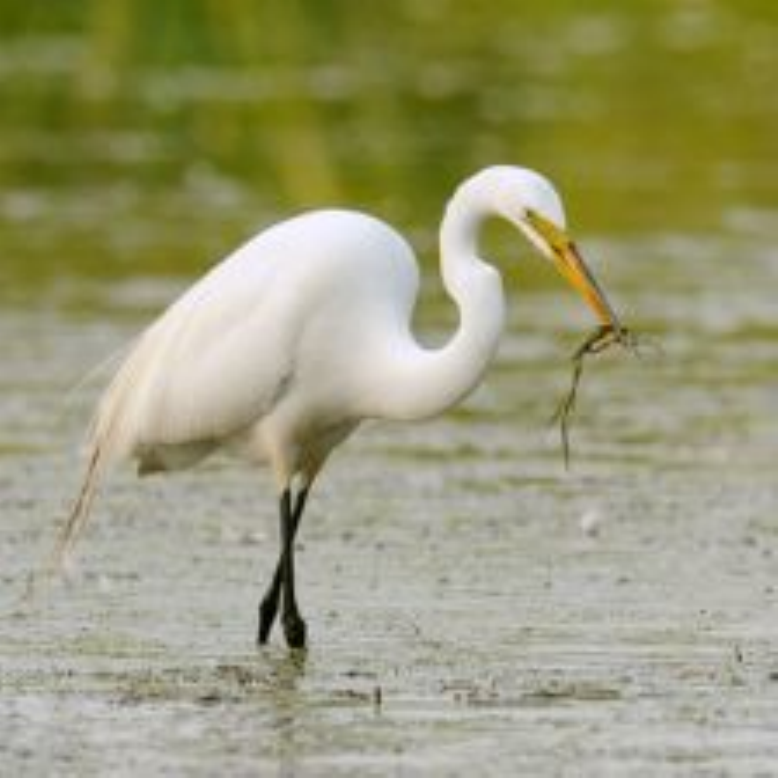}
\includegraphics[width=0.48\linewidth]{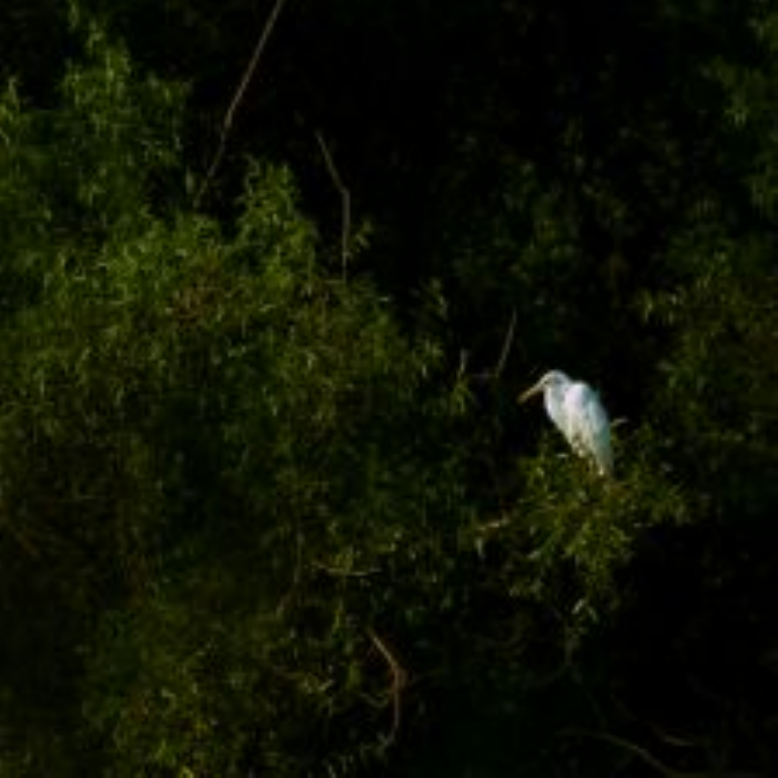}
\caption{American egret}
\end{subfigure} 
\begin{subfigure}[b]{0.48\linewidth}
\centering
\includegraphics[width=0.48\linewidth]{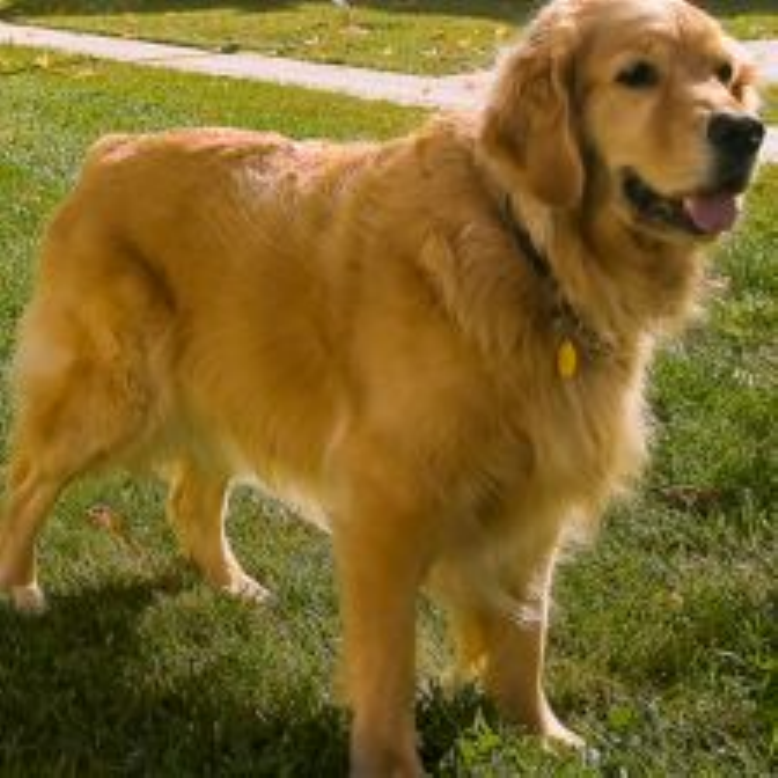}
\includegraphics[width=0.48\linewidth]{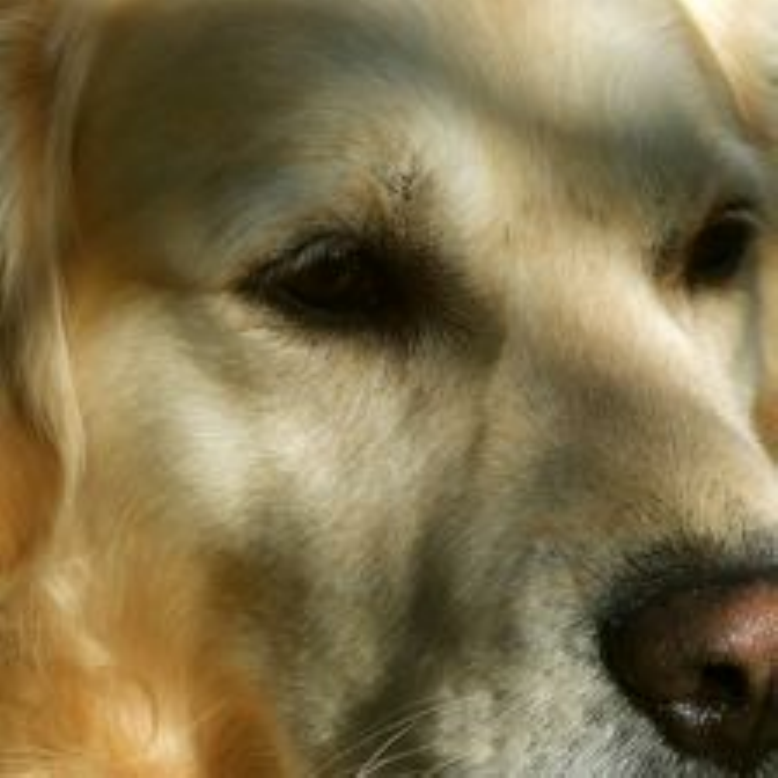}
\caption{Golden retriever}
\end{subfigure}\\
\vskip 0.1in
\begin{subfigure}[b]{0.48\linewidth}
\centering
\includegraphics[width=0.48\linewidth]{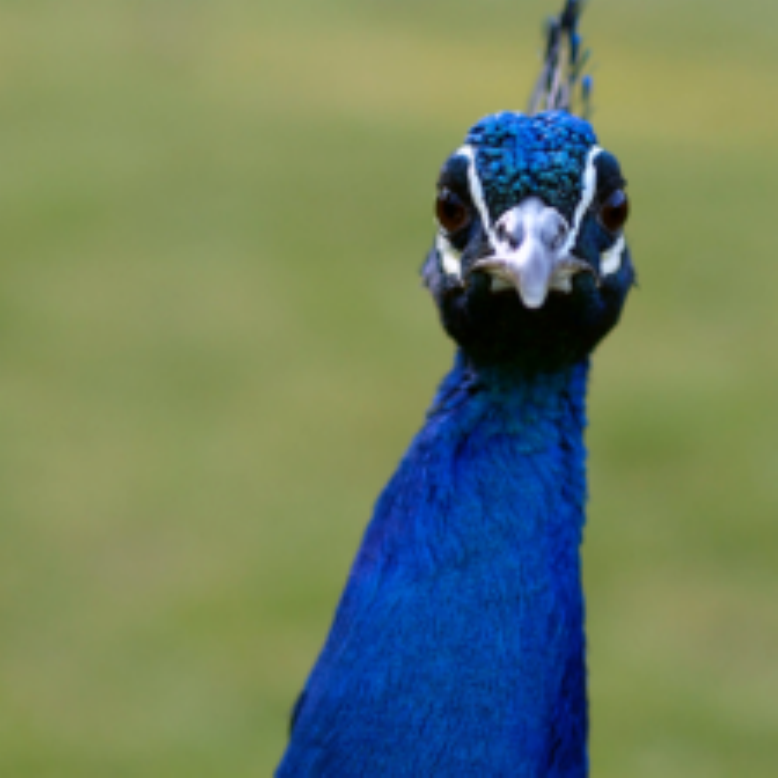}
\includegraphics[width=0.48\linewidth]{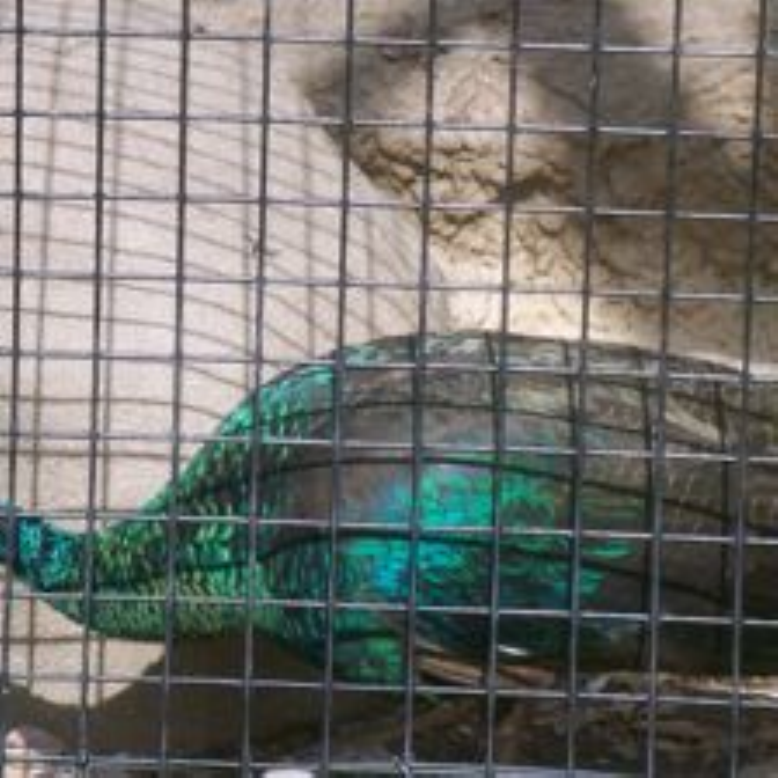}
\caption{Peacock}
\end{subfigure} 
\begin{subfigure}[b]{0.48\linewidth}
\centering
\includegraphics[width=0.48\linewidth]{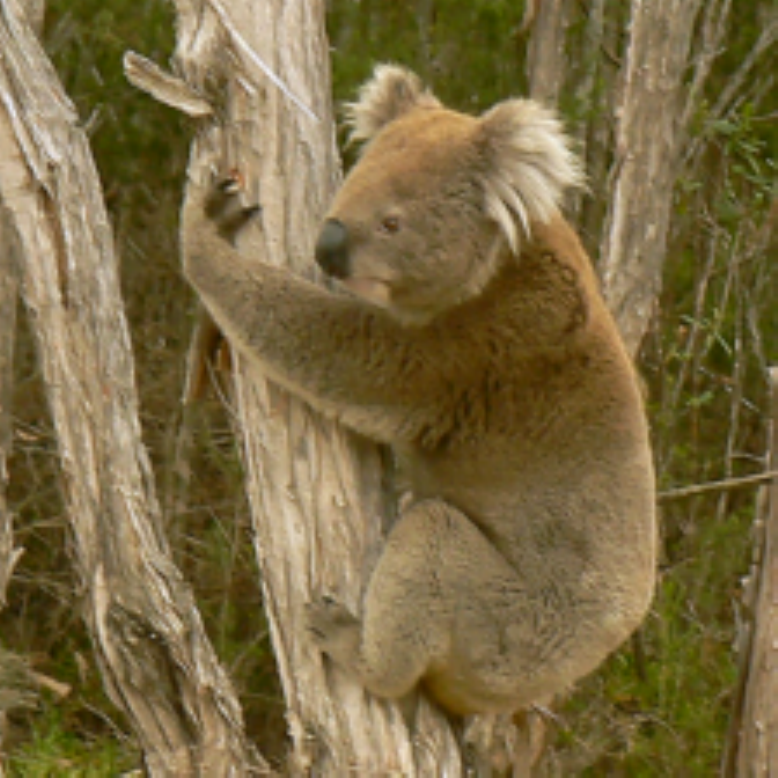}
\includegraphics[width=0.48\linewidth]{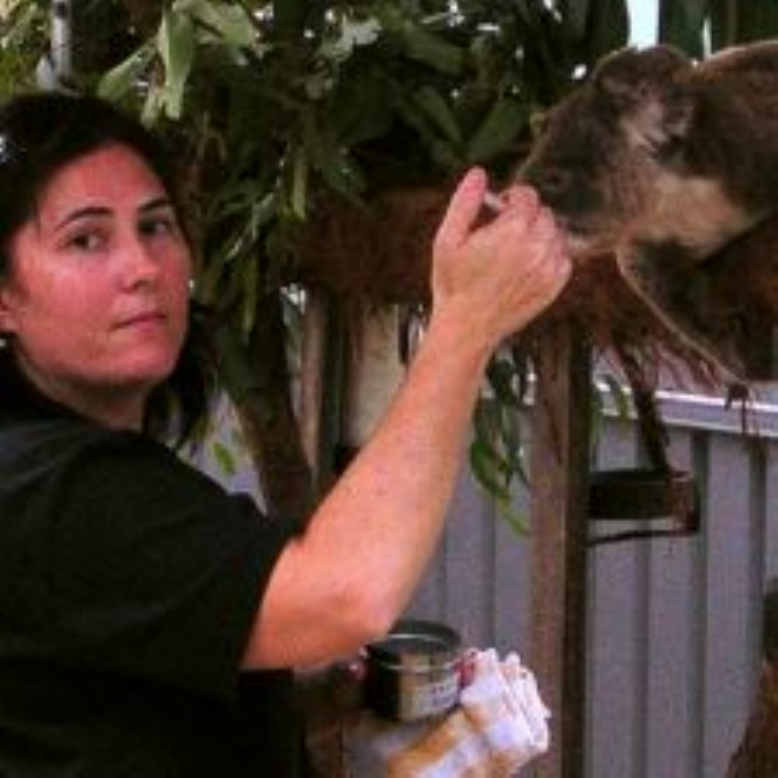}
\caption{Koala}
\end{subfigure} 
\vspace{-3mm}
\caption{ImageNet samples that activate the blocks differently with an equal scale parameter ($\mathcal{S}=0.6$). In each object class, the left ones activate 19 blocks of the network, whereas the right ones activate 23 blocks.}
\label{fig:qualitative_1}
\vspace{-1mm}
\end{figure}

In Figure \ref{fig:usage_map}, there is green, blue area that represents the block usage is about 0.2$\sim$0.8. These blocks are dynamically opened or closed depending on the input image. These blocks may contain minor but detailed features for hard samples. Figure \ref{fig:qualitative_1} is the examples of pair of samples that induce the model to activate blocks differently during inference under given $\mathcal{S}=0.6$. In the Figure, the pair of samples look very different visually. The left ones, which use the minimum number of blocks have very distinctive and remarkable features. Whereas the samples on the right, which need the maximum number of blocks, are hard samples that have too small object (a), too large object (b), too noisy (c), or interrupted by other object (d).

\begin{figure}[t]
\centering
\begin{subfigure}[b]{0.32\linewidth}
\centering
\includegraphics[width=\linewidth]{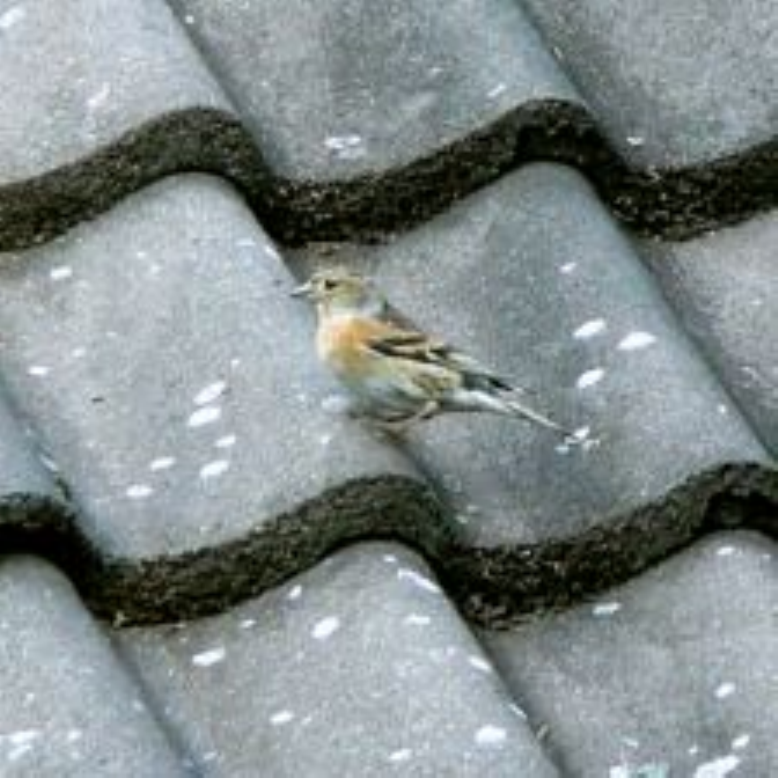}
\caption{Brambling\\\hspace{\textwidth}(Night snake)}
\end{subfigure} 
\begin{subfigure}[b]{0.32\linewidth}
\centering
\includegraphics[width=\linewidth]{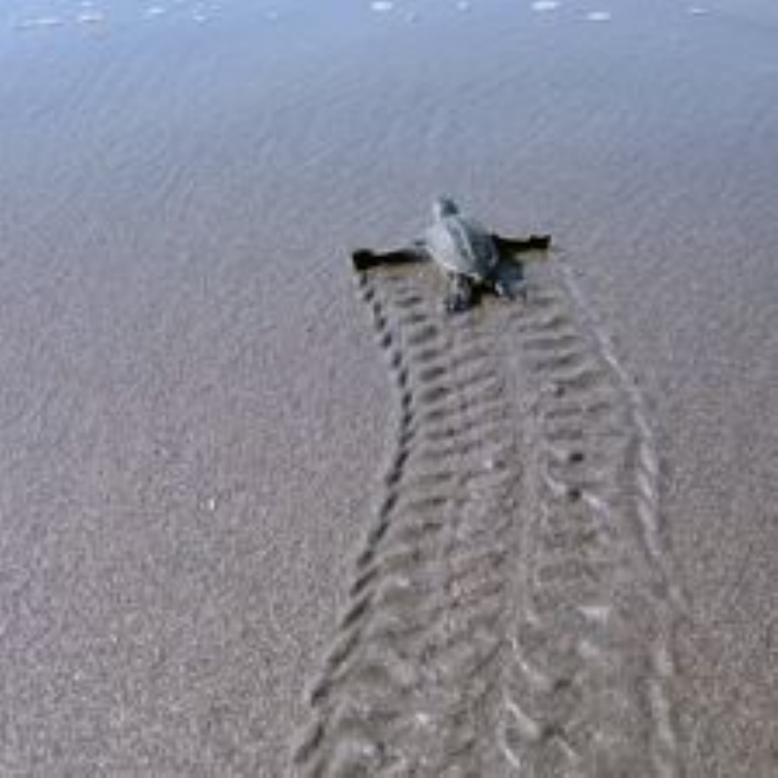}
\caption{Turtle\\\hspace{\textwidth}(Alligator)}
\end{subfigure}
\begin{subfigure}[b]{0.32\linewidth}
\centering
\includegraphics[width=\linewidth]{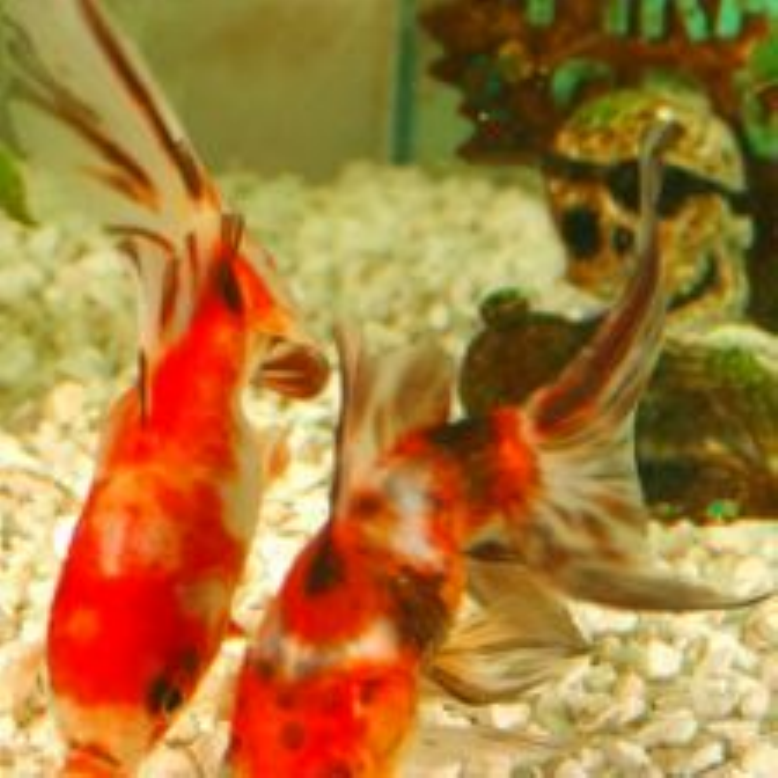}
\caption{Goldfish\\\hspace{\textwidth}(Crayfish)}
\end{subfigure}\\
\vskip 0.1in
\begin{subfigure}[b]{0.32\linewidth}
\centering
\includegraphics[width=\linewidth]{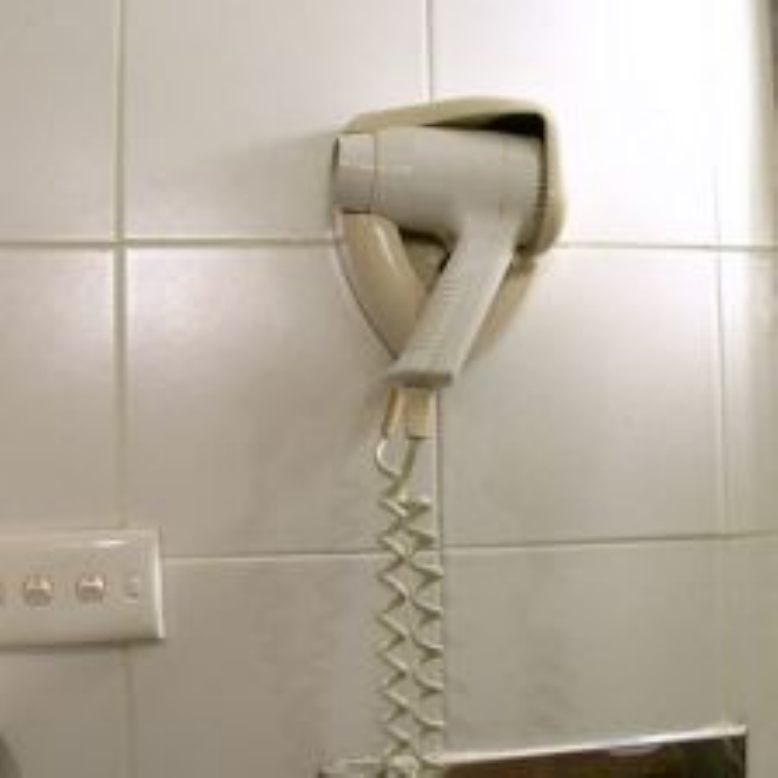}
\caption{Hand blower\\\hspace{\textwidth}(Toilet tissue)}
\end{subfigure}
\begin{subfigure}[b]{0.32\linewidth}
\centering
\includegraphics[width=\linewidth]{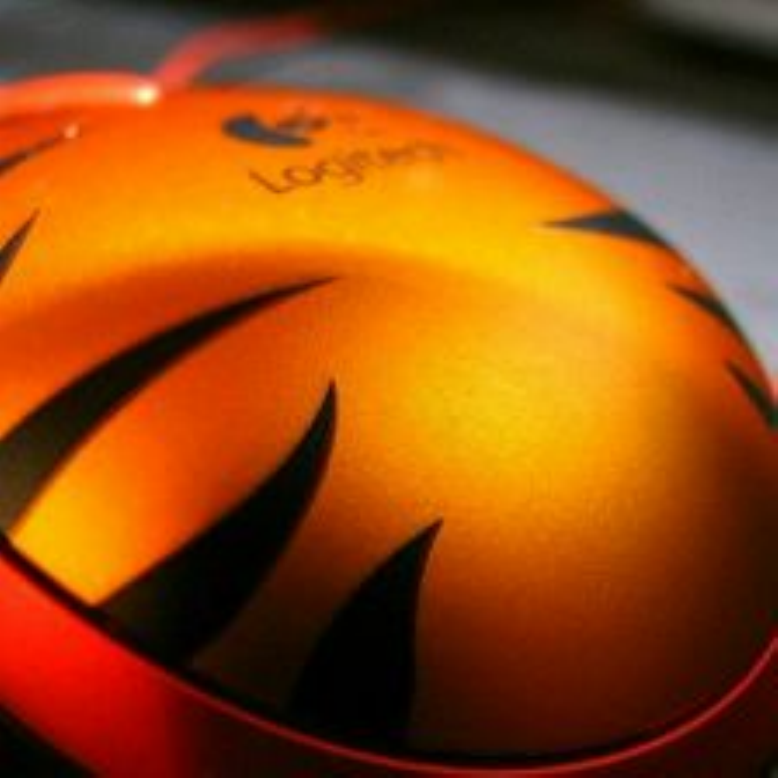}
\caption{Mouse\\\hspace{\textwidth}(Jack O lantern)}
\end{subfigure}
\begin{subfigure}[b]{0.32\linewidth}
\centering
\includegraphics[width=\linewidth]{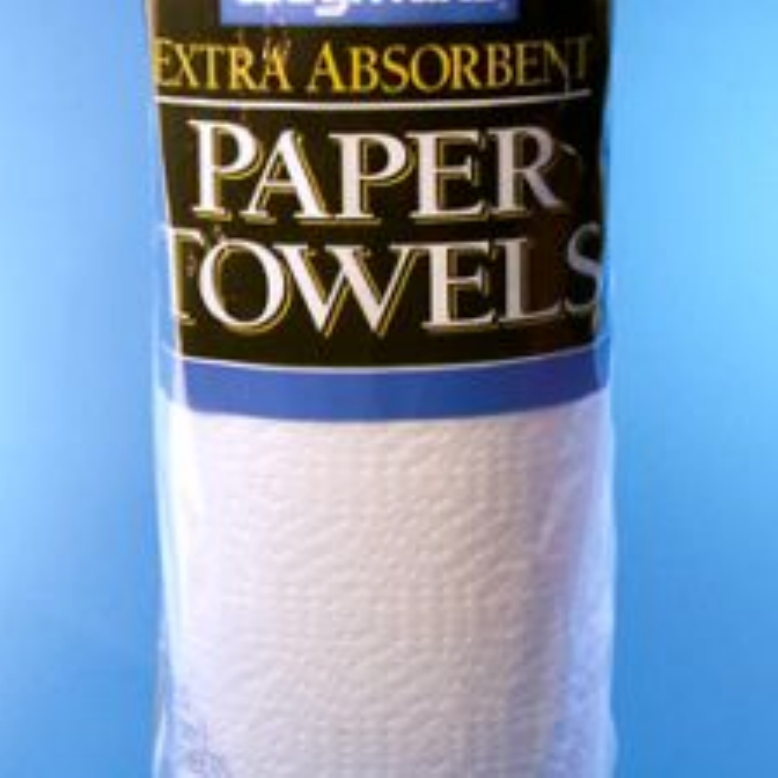}
\caption{Paper towel\\\hspace{\textwidth}(Beer bottle)}
\end{subfigure}
\caption{Samples in ImageNet that was correctly classified for large $\mathcal{S}$ (0.8) but misclassified as $\mathcal{S}$ gets reduced (0.6). 
The misclassified label is written in parentheses. These samples can be regarded as hard samples.}
\label{fig:qualitative_2}
\vspace{-3mm}
\end{figure}

Figure \ref{fig:qualitative_2} shows the examples that are classified correctly under higher $\mathcal{S}$, 
but are misclassified as $\mathcal{S}$ gets reduced.
Those samples are very hard and confusing samples. Without very detailed features, it can be easily fooled by other class.

\begin{table}[t]
\caption{The effect of the CGMs with or without using previous block features as input. The $\beta$ is set to 2.0 for both. For each cells the first low is accuracy and the second low is block usage, and the third low is variance of gate usage. The performance of URNet w/o previous block is slightly worse than URNet in most cases and the \sh{resizability} is also degraded, as the URNet w/o previous block cannot close the block under 21.00.}
\vspace{-4mm}
\begin{center}
 \begin{tabular*}{\linewidth} {@{\extracolsep{\fill} } p{0.1\textwidth} c c c c c}
 \hline
 \multicolumn{6}{c}{CIFAR-10} 							\\ \hline
 $\mathcal{S}$ 							& 0.2	& 0.4	& 0.6	& 0.8	& 1.0		\\ \hline               
  \multirow{3}{*}{  	
  \begin{tabular}[c]{@{}l@{}}URNet
  \end{tabular}}
  										& 92.2	& 93.3	& 93.7	& 93.7	& 93.6		\\ 
  										& 18.08	& 20.86	& 32.02	& 44.37	& 52.19		\\ 
                                        & 0.92	& 1.01	& 1.29	& 1.03	& 0.79		\\ \hline
  \multirow{3}{*}{  	
  \begin{tabular}[c]{@{}l@{}}URNet w/o	\\
  {\small previous block}\end{tabular}}
										& 93.1	& 93.1	& 93.4	& 93.5	& 93.4		\\ 
                                        & 21.00	& 21.00	& 32.00	& 45.00	& 52.00		\\
                                        & 0.00	& 0.00	& 0.00	& 0.00	& 0.00		\\ \hline 
 \end{tabular*}
\label{table:woblock}
\end{center}
\vspace{-4mm}
\end{table}

\subsection{Resizable Range}

\begin{figure}[t]
\centering
\vskip -0.1in
\includegraphics[width=1.\linewidth]{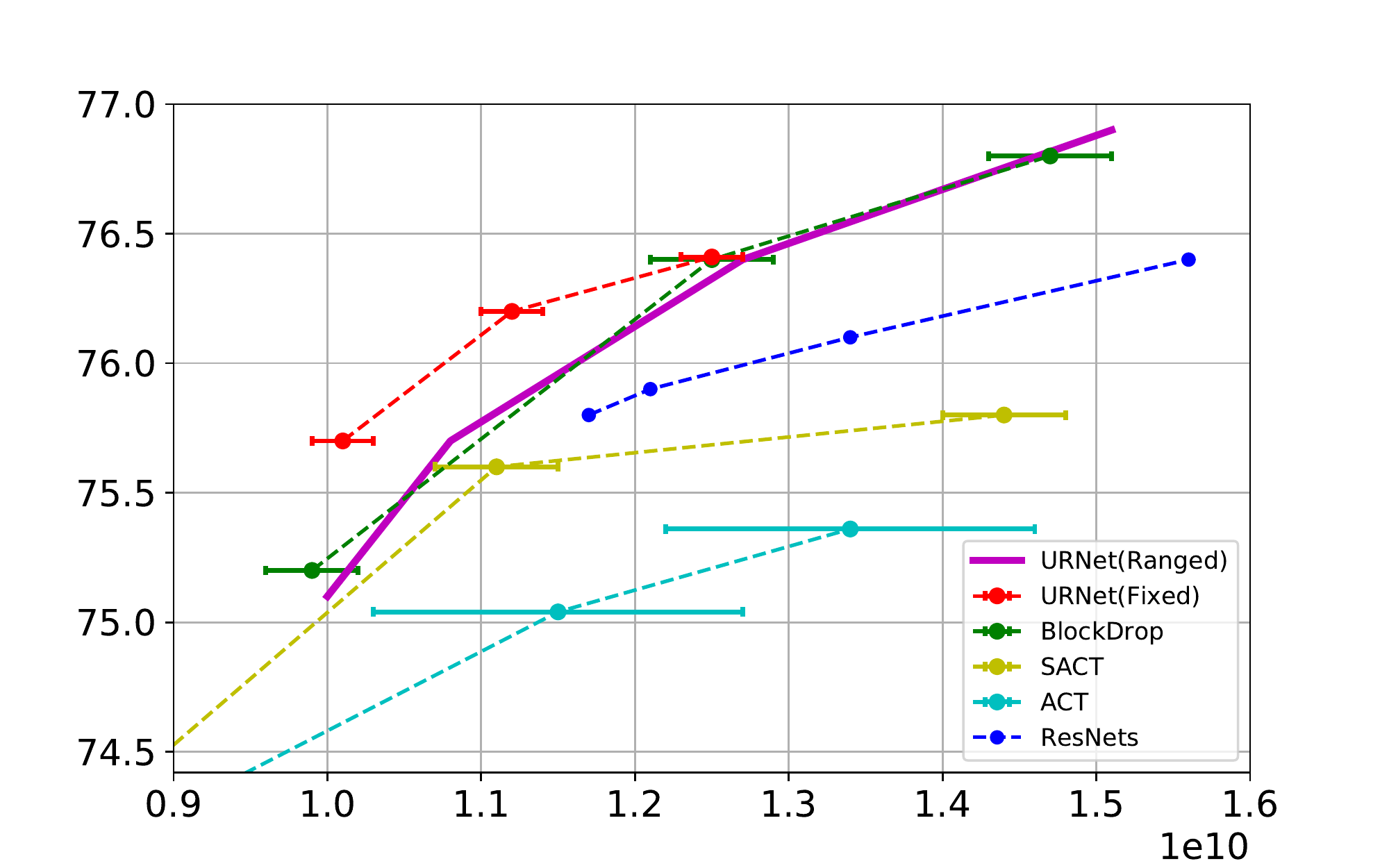}
\vskip -0.05in
\caption{\sh{\textbf{Accuracy vs FLOP}}. This figure compares URNet(Ranged) and URNet(Fixed) on ImageNet with other methods \cite{wu2018blockdrop, sact}. The dot represents one model, and the solid horizontal line represents the standard deviation of one model. URNet(Ranged) represents user resized results at test time \sh{by} one model. Those of ResNet-\{72, 75, 84, 110\} and other results are all brought from \cite{wu2018blockdrop}.}
\label{fig:compression}
\vspace{-5mm}
\end{figure}

Our URNet can obtain accuracy/FLOPs similar to state-of-the-art compression methods, even though ours has additional characteristics of resizability. Therefore, as stated in section \ref{sec:related}, our model does not need to prepare many different-sized networks at once on memory. As stated previously, we have trained $\mathcal{S}$ with $0.2\sim1.0$, but it is hard to satisfy both high performance and large range of $\mathcal{S}$ simultaneously and there exists trade-off between them. In an environment that accepts a more narrow range of $\mathcal{S}$, there is a room to boost performance.

If we train a network with a fixed $\mathcal{S}$ ($\mathcal{S}_{fixed}$), our method can be considered as a static compression method. In this scenario, there is no need to consider the model architecture (number of blocks, kernel size, channel size, etc.) and we just need to set  $\mathcal{S}_{fixed}$ as a desirable size.

While the resizable one (URNet(Ranged)) uses various values of $\mathcal{S}$ during training, the fixed scale URNet(Fixed) uses only a small fraction of entire range of $\mathcal{S}$, so there may be the case where only a few blocks are selected to use from the beginning of the training, rather than considering various blocks. To prevent this, the scale parameter $\mathcal{S}_{fixed}$ is initially set to 1, and then gradually reduced to a desirable size. This is called \textit{Scale Annealing} and $\mathcal{S}_{fixed}$ is decayed with the cosine annealing schedule~\cite{loshchilov2016sgdr} for specific epochs. 
In addition, to keep the ability of selectively using blocks, the Gaussian noise is added so $\mathcal{S}_{fixed}$ is sampled from $\mathcal{N}(\mathcal{S}_{fixed}, \sigma^2)$ but restricted not to exceeds 1.

Figure \ref{fig:compression} shows the accuracy versus FLOPs of URNet and other compression methods on ImageNet. The solid horizontal line in the figure represents the standard deviation of FLOPs of one model at test time. Note that the URNet(Ranged) is just one model, and can be resized according to user's demand, that others cannot.
The URNet(Fixed) is trained with $\mathcal{S}_{fixed} =$ 0.5, 0.6 and 0.7, and the Gaussian noise with $\sigma=0.1$ is added to $\mathcal{S}_{fixed}$ at training time. 5 epochs of scale annealing is applied. Our URNet(Ranged) performs almost equal to BlockDrop, and URNet(Fixed) performs better than that.

\section{Conclusion}
We showed that our User-Resizable Residual Networks (URNet) can resize itself as a response to the demand of a user, at any inference time. \sh{Experimental results show} that our URNet can change its computational cost without severe accuracy degradation. Unlike other methods, it does not need a reinforcement learning algorithm to use a binary step function and can be trained in a \sh{simple supervised manner.} Our method can be applied to any ResNet-based network with very little ($<$0.1\%) additional computational burden. Using our method, the user of a network can dynamically balance the number of requests executed per time, by dynamically adjusting the amount of resources per request.


{\small
\bibliographystyle{ieee}
\bibliography{egbib}
}

\onecolumn
\clearpage

\noindent\Large{\textbf{Supplementary Material}}
\setcounter{section}{0}

\renewcommand\thesection{\Alph{section}}

\normalsize

\section{Experiment Details}

\subsection{CIFAR Datasets}
We use ADAM as an optimizer for CIFAR datasets with a batch size of 256. The learning rate is decayed from $ 10 ^ {- 3} $ to $ 10 ^ {- 5} $ at 300 and 400 epoch of total 500 epochs. And we train CGM only for \nj{the} first 100 epochs, and train CGM and ResNet jointly for the remaining 400 epochs.\\


\subsection{ImageNet Dataset}
On ImageNet, we use ADAM for CGMs and momentum SGD for ResNet with momentum 0.9 and weight decay $10^{-4}$. The first 10 of total 20 epochs is trained with learning rate $10^{-3}$ and the remaining each 5 epochs use the learning rate $10^{-4}$ and $10^{-5}$, respectively. Similar to CIFAR, we train CGM only for \nj{the} first 5 epochs, and train CGM and ResNet jointly for the remaining 15 epochs with a batch size of 340.
\\

\subsection{External Gating Network}
The \textit{External network} in Table 1 of the \nj{main} paper is a separate external network for block gates, similar to the Policy Network used for CIFAR datasets in \cite{wu2018blockdrop}.
This network is a small ResNet with three residual blocks (ResNet-8), which takes depth-wise concatenation of the input image and the scale parameter $S$ repeated in the same spatial size. Then, it outputs the gates for the blocks of the base network with the last fully connected layer.
The gates use a sigmoid function with gate training probability $p$ and a binary step function with probability $1-p$ as an activation function, \nj{in the same way as} the Conditional Gating Module (CGM) of URNet.

\section{Quantitative Result}
Table \ref{table:detail} shows \nj{more detailed results} of CIFAR datasets and ImageNet, with variance of blocks and FLOPs.

\begin{table*}[h]
\centering
\caption{The expanded result of Table 1 and Table 2 in the \nj{main} paper.}
\begin{subtable}[t]{0.99\textwidth}
\centering
 \begin{tabular*}{0.99\textwidth} {@{\extracolsep{\fill} } p{0.1\textwidth}  c c c c c }
 \hline
 								\multicolumn{6}{c}{CIFAR-10} 			   															\\ 
 $\mathcal{S}$ 					&  0.2				&  0.3				&  0.4				&  0.5				& 0.6  				\\ \hline 
  \multirow{3}{*}{  	
  \begin{tabular}[c]{@{}l@{}}Accuracy\\Block usage\\FLOPs(E+8)\end{tabular}}
								& 92.2				& 93.2				& 93.3				& 93.7 				& 93.7	 			\\ 
                                & 18.08$\pm$0.92	& 18.70$\pm$0.94	& 20.86$\pm$1.01	& 26.63$\pm$1.18	& 32.02$\pm$1.29	\\
 								& 1.67$\pm$0.09		& 1.73$\pm$0.09		& 1.93$\pm$0.10		& 2.48$\pm$0.11		& 2.99$\pm$0.12		\\ \hline\hline
 $\mathcal{S}$ 					& 0.7				& 0.8				& 0.9				& 1.0				&   				\\ \hline 
  \multirow{3}{*}{  	
  \begin{tabular}[c]{@{}l@{}}Accuracy\\Block usage\\FLOPs(E+8)\end{tabular}}
								& 93.6				& 93.7				& 93.7				& 93.6 				& 		 			\\ 
                                & 38.56$\pm$1.48	& 44.37$\pm$1.03	& 49.98$\pm$1.08	& 52.19$\pm$0.79	& 					\\
 								& 3.60$\pm$0.14		& 4.15$\pm$0.10		& 4.68$\pm$0.10		& 4.89$\pm$0.07		& 					\\ \hline
 \end{tabular*}
 \caption{CIFAR-10, URNet-110, $\beta=2.0$, $p=0.1$, total 54 blocks.}
\end{subtable}

\vspace{0.7cm}
\begin{subtable}[t]{0.99\textwidth}
 \centering
 \begin{tabular*}{0.99\textwidth} {@{\extracolsep{\fill} } p{0.1\textwidth}  c c c c c }
 \hline
 								\multicolumn{6}{c}{CIFAR-100} 			   						\\ 
 $\mathcal{S}$ 					&  0.2				&  0.3				&  0.4				&  0.5				& 0.6  				\\ \hline
  \multirow{3}{*}{  	
  \begin{tabular}[c]{@{}l@{}}Accuracy\\Block usage\\FLOPs(E+8)\end{tabular}}
								& 70.7				& 71.1				& 71.5				& 71.8 				& 72.3	 			\\ 
                                & 28.10$\pm$0.54	& 28.28$\pm$0.65	& 28.57$\pm$0.80	& 29.00$\pm$0.97	& 31.99$\pm$1.29	\\
 								& 2.62$\pm$0.05		& 2.63$\pm$0.06		& 2.66$\pm$0.08		& 2.70$\pm$0.09		& 2.98$\pm$0.12		\\ \hline\hline
 $\mathcal{S}$ 					& 0.7				& 0.8				& 0.9				& 1.0				&   				\\ \hline                  
  \multirow{3}{*}{  	
  \begin{tabular}[c]{@{}l@{}}Accuracy\\Block usage\\FLOPs(E+8)\end{tabular}}
								& 72.6				& 73.0				& 72.8				& 72.8 				& 		 			\\ 
                                & 36.58$\pm$1.54	& 44.61$\pm$1.78	& 49.06$\pm$0.96	& 49.41$\pm$0.80	& 					\\
 								& 3.42$\pm$0.15		& 4.17$\pm$0.17		& 4.59$\pm$0.09		& 4.63$\pm$0.08		& 					\\ \hline
 \end{tabular*}
 \caption{CIFAR-100, URNet-110, $\beta=2.0$, $p=0.1$, total 54 blocks.}
\end{subtable}

\vspace{0.7cm}
\begin{subtable}[t]{0.99\textwidth}
 \centering
 \begin{tabular*}{0.99\textwidth} {@{\extracolsep{\fill} } p{0.15\textwidth}  c c c c c }
 \hline
 								\multicolumn{6}{c}{ImageNet} 			   															\\ 
 $\mathcal{S}$ 					&  0.2				&  0.3				&  0.4				&  0.5				& 0.6  				\\ \hline
  \multirow{3}{*}{  	
  \begin{tabular}[c]{@{}l@{}}Accuracy\\Block usage\\FLOPs(E+8)\end{tabular}}
								& 74.0				& 74.5				& 74.9				& 75.3 				& 75.7	 			\\ 
                                & 18.77$\pm$0.70	& 19.17$\pm$0.83	& 19.76$\pm$0.86	& 20.60$\pm$0.75	& 22.01$\pm$1.01	\\
 								& 0.94$\pm$0.03		& 0.96$\pm$0.04		& 0.98$\pm$0.04		& 1.02$\pm$0.03		& 1.08$\pm$0.04  	\\ \hline\hline
 $\mathcal{S}$ 					& 0.7				& 0.8				& 0.9				& 1.0				&   				\\ \hline
  \multirow{3}{*}{  	
  \begin{tabular}[c]{@{}l@{}}Accuracy\\Block usage\\FLOPs(E+8)\end{tabular}}
								& 76.2				& 76.4				& 76.8				& 76.9 				& 		 			\\ 
                                & 25.13$\pm$0.82	& 26.94$\pm$0.40	& 31.13$\pm$0.43	& 32.00$\pm$0.02	& 					\\
 								& 1.22$\pm$0.04		& 1.30$\pm$0.02		& 1.48$\pm$0.02		& 1.52$\pm$0.00		& 					\\ \hline
 \end{tabular*}
 \caption{ImageNet, URNet-101, $\beta=4.0$, $p=0.1$, total 33 blocks.}
\end{subtable}
\label{table:detail}
\end{table*}


\section{URNet on Object Detection}
We have \nj{applied} the URNet method on to the object detection task. We \nj{applied it} on the Faster-RCNN\cite{ren2015faster}, trained and tested on the Pascal VOC 2007~\cite{pascal-voc-2007} dataset. First, we implemented the Faster-RCNN with ResNet-101 with \nj{the} pretraiend ImageNet weight, and \nj{achieved} mAP$=74.96$. And we trained the URNet on this object detector network with 1 epoch of CGMs training, and 12 epochs of joint training. The CGMs are attached to the blocks between \nj{the} 4th block and \nj{the} 30th block, \nj{i.e. a total of 27 CGMs}. As shown in Table \ref{table:voc}, our method \nj{is} resizable on object \nj{detection} task well, and achieves more performance gain with less computation.

\begin{table}
\centering
 \caption{Object \nj{detection} result on the Pascal VOC 2007, $\beta=2.0$, $p=0.1$, total 33 blocks. Each cells the first row is mAP, and the second row is the number of block usage.}
 \begin{tabular*}{0.99\textwidth} {@{\extracolsep{\fill} } p{0.15\textwidth}  c c c c c }
 \hline
 								\multicolumn{6}{c}{Pascal VOC 2007} 			   															\\ 
 $\mathcal{S}$ 					&  0.2				&  0.4				&  0.6				&  0.8				& 1.0  				\\ \hline
  \multirow{2}{*}{  	
  \begin{tabular}[c]{@{}l@{}}Faster-RCNN\\(ResNet-101)\end{tabular}}
                                & -					& -					& -					& -					& 74.96					\\
                                & -					& -					& -					& -	 				& 33					\\ \hline
  \multirow{2}{*}{  	
  \begin{tabular}[c]{@{}l@{}}URNet\\(on Faster-RCNN)\end{tabular}}
								& 72.84				& 73.67				& 74.34				& 75.26				& 75.40	 			\\ 
                                & 19.65$\pm$0.51	& 20.11$\pm$0.32	& 22.91$\pm$0.40	& 28.01$\pm$0.74	& 30.00$\pm$0.00	\\ \hline
 \end{tabular*}
 \label{table:voc}
\end{table}

\end{document}